\documentclass{article}

\usepackage[preprint]{neurips_2024}
\usepackage[pdftex]{graphicx}
\usepackage[dvipsnames]{xcolor}
\usepackage[utf8]{inputenc} % allow utf-8 input
\usepackage[T1]{fontenc}    % use 8-bit T1 fonts
\usepackage{hyperref}       % hyperlinks
\usepackage{url}            % simple URL typesetting
\usepackage{booktabs}       % professional-quality tables
\usepackage{amsfonts}       % blackboard math symbols
\usepackage{nicefrac}       % compact symbols for 1/2, etc.
\usepackage{microtype}      % microtypography
\usepackage{xcolor}         % colors
\usepackage{float}
\usepackage[raster,most]{tcolorbox}
\setcitestyle{numbers,square}
\usepackage{enumitem}
\usepackage{multirow}
\usepackage{multicol}
\usepackage{tikz}
\usepackage[most]{tcolorbox}
\usetikzlibrary{arrows.meta, positioning}
\usepackage{hyperref}

\title{GOAT-SLM: A Spoken Language Model with Paralinguistic and Speaker Characteristic Awareness}

\definecolor{gray}{HTML}{777777}
\definecolor{}{HTML}{00008B}

\author{
    Hongjie Chen$^{\dag}$ \And
    Zehan Li$^{\dag}$ \And
    Yaodong Song$^{\dag}$ \And 
    Wenming Deng$^{\dag}$ \And
    Yitong Yao$^{\dag}$ \And
    Yuxin Zhang$^{\dag}$ \And
    Hang Lv$^{\dag}$ \And
    Xuechao Zhu \And
    Jian Kang \And
    Jie Lian \And
    Jie Li \And
    Chao Wang \And
    Shuangyong Song \And
    Yongxiang Li \And
    Zhongjiang He \And
    Xuelong Li$^{*}$ \\
    \\
    Institute of Artificial Intelligence (TeleAI), China Telecom, China
}

\begin{document}

\maketitle
\renewcommand{\thefootnote}{}
\footnotetext{$^\dag$ Equal contribution.}
\footnotetext{$^*$ Corresponding author. Email: \href{mailto:xuelong_li@ieee.org}{xuelong\_li@ieee.org}}
\setcounter{footnote}{0}

\begin{abstract}
Recent advances in end-to-end spoken language models (SLMs) have significantly improved the ability of AI systems to engage in natural spoken interactions. However, most existing models treat speech merely as a vehicle for linguistic content, often overlooking the rich paralinguistic and speaker characteristic cues embedded in human speech, such as dialect, age, emotion, and non-speech vocalizations.
In this work, we introduce GOAT-SLM, a novel spoken language model with paralinguistic and speaker characteristic awareness, designed to extend spoken language modeling beyond text semantics. GOAT-SLM adopts a dual-modality head architecture that decouples linguistic modeling from acoustic realization, enabling robust language understanding while supporting expressive and adaptive speech generation.
To enhance model efficiency and versatility, we propose a modular, staged training strategy that progressively aligns linguistic, paralinguistic, and speaker characteristic information using large-scale speech-text corpora. Experimental results on TELEVAL, a multi-dimensional evaluation benchmark, demonstrate that GOAT-SLM achieves well-balanced performance across both semantic and non-semantic tasks, and outperforms existing open-source models in handling emotion, dialectal variation, and age-sensitive interactions.
This work highlights the importance of modeling beyond linguistic content and advances the development of more natural, adaptive, and socially aware spoken language systems.
\end{abstract}

\section{Introduction}
In recent years, end-to-end spoken language models (SLMs)~\cite{SLMSurvey} have made rapid progress~\cite{speechgpt,moshi,speechgpt2,llama-omni,freeze-omni,slam-omni,glm4voice,minicpm-o,baichuan-omni15,salmonn-omni,minmo,Qwen25-omni,step-audio,step-audio-aqaa,llama-omni2,deeptalk}. However, most existing models still regard speech primarily as a carrier of linguistic content, often overlooking the rich paralinguistic and speaker characteristic cues embedded within the audio signal. These cues—including dialect, age, emotion, and non-speech vocal (NSV) signals such as coughing—are essential for enabling more natural, adaptive, and empathetic human-computer interactions~\cite{boss}. To bridge this gap, we propose a spoken language model that can perceive and respond to such non-linguistic vocal information, enhancing interaction quality through paralinguistic and speaker characteristic awareness.

In terms of model construction, SLMs typically follow two main paradigms. The first is the speech-native approach (e.g., Moshi~\cite{moshi}, SpeechGPT~\cite{speechgpt}, SpeechGPT2~\cite{speechgpt2}, Viola~\cite{voila2025}, Baichuan-Omni-1.5~\cite{baichuan-omni15}), which directly applies large language model (LLM) training paradigms to large-scale tokenized speech corpora. The second is the modality-alignment approach (e.g., Freeze-Omni~\cite{freeze-omni}, Salmonn-Omni~\cite{salmonn-omni}, MinMo~\cite{minmo}, Kimi-Audio~\cite{kimi-audio}), which leverages a strong LLM core and integrates speech input/output modules via cross-modal alignment. In this work, we adopt the modality-alignment framework as our foundation, due to its computational efficiency and ability to reuse pretrained speech and text models.

We introduce GOAT-SLM, a novel spoken language model designed around four core principles:  
\textbf{G}eneration-oriented dual-modality design,  
\textbf{O}rchestrated modular training,  
\textbf{A}wareness of paralinguistic and speaker characteristics, and  
\textbf{T}ext-intelligence preservation.
As shown in Figure~\ref{fig:framework}, GOAT-SLM adopts a dual-modality head architecture where the shared lower layers of a pretrained LLM act as a semantic core. This core branches into two modality-specific heads: one for text generation and another for speech token generation. The speech head is initialized from the text head to promote parameter sharing and knowledge transfer. This bifurcated architecture brings several key advantages. First, by decoupling linguistic modeling from acoustic realization, it preserves the LLM’s core reasoning and understanding abilities. Second, it enables high-fidelity and expressive speech synthesis by leveraging rich semantic representations. Third, the architecture supports flexible task-specific training. For example, training on large-scale ASR or TTS datasets yields strong TTS performance. Alternatively, training on paired speech-to-speech QA data enables spoken question answering. In fact, we extend this framework to implement GOAT-TTS~\cite{goat-tts}, which is used to generate synthetic speech-to-speech dialogue data for training GOAT-SLM.

For training, we introduce an orchestrated modular strategy composed of three stages:
\begin{itemize}
\item Instruction tuning:  We inject attributes such as dialect, age, and emotion into user instructions, allowing the LLM to recognize and respond to fine-grained vocal cues. This equips LLM with the ability to generate more empathetic and contextually appropriate responses.

   \item Modality alignment training:  
   Using repetition and continuation strategies, we align speech and text modalities while freezing the LLM backbone. This enables speech-text grounding without degrading the LLM’s pretrained capabilities.

   \item High-fidelity speech generation optimization:  
   We refine the speech head to enhance naturalness, intelligibility, and expressiveness. Throughout all stages, the textual intelligence of the LLM is preserved, ensuring strong general reasoning and instruction-following capabilities.
\end{itemize}

To evaluate the model’s ability to perceive and respond to paralinguistic and speaker characteristic signals, we perform evaluation on a comprehensive dialogue evaluation framework, TELEVAL~\cite{Benchmark}.  
The evaluation results show that GOAT-SLM demonstrates strong performance across multiple dimensions, including general knowledge question answering, acoustic robustness, empathetic dialogue, dialectal interaction, and age-sensitive response generation.  
Dialogue samples in inference are available at our demo site$^{1}$\footnotetext{$^{1}$ https://tele-ai.github.io/GOAT-SLM.github.io/}.

\section{Related Work}
Recent advances in large language models (LLMs) have significantly shaped the development of spoken language models (SLMs), which aim to handle spoken dialogues in an end-to-end fashion. While early work focused on bridging language and speech modalities for semantic comprehension, recent studies have emphasized the need to preserve core language intelligence while incorporating audio-specific capabilities into dialogue systems. In this section, we review related work along these two lines of research and outline how our approach builds upon and extends them.

\subsection{Preserving Linguistic Intelligence in Spoken Language Models}

The integration of LLMs into SLMs leverages the strong language understanding and generation capabilities acquired from large-scale text corpora. However, adapting LLMs to speech tasks often risks catastrophic forgetting, where the model's language competence deteriorates after fine-tuning on multimodal data. Addressing this challenge has become a central concern in recent research.

One prevalent strategy is curriculum-style learning, where SLMs undergo continual pre-training on large-scale multimodal datasets before being fine-tuned with lightweight instruction tuning. This staged training paradigm enables the model to acquire audio comprehension capabilities while mitigating degradation of its linguistic intelligence. For instance, SpeechGPT~\cite{salmonn-omni} and Moshi~\cite{moshi} both adopt this approach, demonstrating that continual multimodal pre-training followed by targeted instruction tuning effectively preserves the model's language capabilities.

Another effective strategy is to freeze the core parameters of the LLM and introduce a small set of learnable parameters through auxiliary mechanisms, minimizing the risk of interfering with the language backbone. For example, MinMo~\cite{minmo} adopts LoRA-based fine-tuning during speech-to-text alignment training to maintain the stability of the LLM core. Freeze-Omni~\cite{freeze-omni} applies prefix tuning for modality adaptation. DeepTalk~\cite{deeptalk} introduces audio-specific expert modules within a mixture-of-experts (MoE) architecture to handle speech-specific information.

Our work aligns with the second strategy through a multi-branch modeling framework designed for joint text and speech generation. Specifically, we introduce a parallel speech generation branch that allows the model to handle both modalities simultaneously without altering the core LLM architecture. Notably, Kimi-Audio~\cite{kimi-audio} shares a similar architecture and was released around the same time as our GOAT-TTS model~\cite{goat-tts}. However, unlike Kimi-Audio, which employs a randomly initialized speech branch, our approach initializes the speech branch with the upper Transformer layers of the pretrained LLM. This design enables tighter parameter sharing and better preserves the LLM’s linguistic competence while extending its capabilities for speech-aware dialogue generation.

\subsection{Modeling Audio-Specific Intelligence in Large Audio Language Models}

While much attention has been given to preserving linguistic intelligence, another essential aspect of SLM design lies in modeling audio-specific intelligence—the ability to interpret and respond to non-linguistic vocal cues that carry contextual and affective information. Recent large audio language models have made notable progress in this direction, demonstrating the capacity to recognize various audio-based attributes such as emotion, language/dialect,  speaker gender and speaker age~\cite{qwen-audio, baichuan-audio, step-audio, kimi-audio}. 
However, these efforts primarily focus on the recognition or description of the user's speech inputs, lacking the ability to autonomously adjust responses based on the detected signals. 
As a result, most models remain reactive and prompt-dependent, limiting their ability to engage in more natural, context-aware spoken interactions.

A smaller body of work has begun exploring responsive intelligence—the ability of a model to dynamically react to paralinguistic and speaker-related cues in user speech. For example, recent models~\cite{glm4voice, step-audio-aqaa, Qwen25-omni, baichuan-omni15} introduce emotion-aware speech generation, allowing the output style to shift according to user-specified emotional states. However, such approaches often rely on explicit control signals rather than autonomous interpretation and context-aware response.

Our approach advances this direction by enabling proactive interpretation and adaptive generation based on paralinguistic and speaker characteristic awareness. The model can detect cues such as dialectal features, emotional states, and non-speech vocalizations, and autonomously adjust its responses without requiring explicit instructions. For instance, it may reply in a matching dialect when the user speaks a regional variant or express empathy when signals like coughing or sighing are detected.

\begin{figure}[!htbp]
	\centering
	\includegraphics[width=1\linewidth]{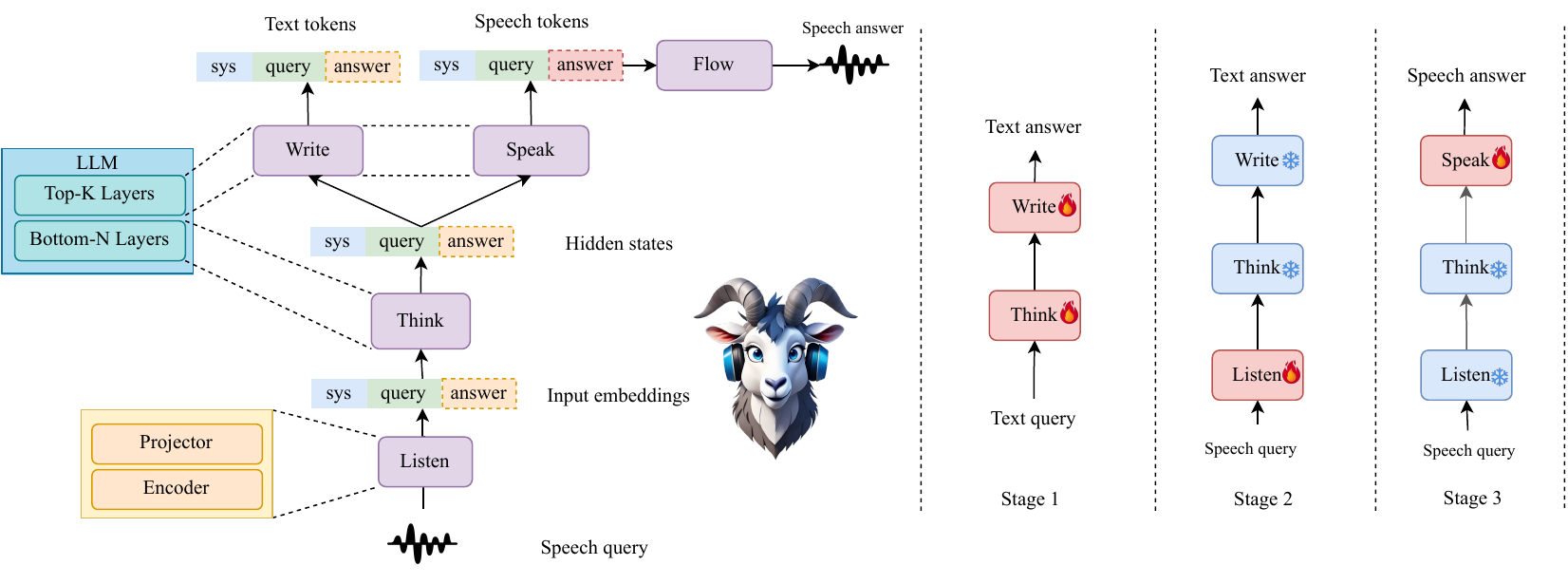} 
    % \hspace{0.02\linewidth}
        \vspace{-1em}
	\caption{Framework and staged training process. Stage 1: Instruction tuning. Stage 2: speech-text alignment training. Stage 3: high-fidelity expressive speech generation training.}
	\label{fig:framework}
\end{figure}

\section{Architecture}\label{sec:Architecture}

Our model architecture, illustrated in Figure~\ref{fig:framework}, consists of five functional modules—\textbf{Listen}, \textbf{Think}, \textbf{Write}, \textbf{Speak}, and \textbf{Flow-Matching}—designed to support unified speech and language understanding and generation. These modules work together to enable seamless spoken dialogue interaction with paralinguistic and speaker characteristic awareness.

The \textbf{Listen} module comprises a speech encoder and a projector, which together transform input speech into latent representations aligned with the LLM’s embedding space. This enables speech inputs to be processed in the same token space as text, facilitating cross-modal semantic integration.

The \textbf{Think} module consists of the bottom $N$ layers of the LLM and serves as a shared semantic reasoning core. The \textbf{Write} module utilizes the top $K$ layers of the LLM to generate textual responses, forming a \textit{Think--Write} pathway for standard text-based interaction. The \textbf{Speak} module reuses the same top $K$ layers but adapts the output layer to predict speech tokens, enabling a \textit{Think--Speak} pathway for speech generation.

The \textbf{Flow-Matching} module serves as a speech decoder, converting predicted speech token sequences into acoustic features (e.g., spectrograms), which are then synthesized into waveforms by a spectrogram-to-wave vocoder.

In our implementation, we adopt Whisper-small$^{1}$\footnotetext{$^1$https://huggingface.co/openai/whisper-small} as the speech encoder, followed by a two-layer convolutional neural network and a two-layer Transformer as the projector in the \textbf{Listen} module. For the language backbone, we use TeleChat2-7B~\cite{telechat}, where the bottom 15 layers form the \textbf{Think} module, and the top 15 layers are shared by the \textbf{Write} and \textbf{Speak} modules for generating text and speech tokens, respectively.

\section{Training}
We adopt a three-stage training strategy to progressively equip GOAT-SLM with robust speech-text alignment, paralinguistic and speaker characteristic awareness, and high-fidelity expressive speech generation, as illustrated in Figure~\ref{fig:framework}. The paralinguistic and speaker characteristic information modeled by GOAT-SLM is shown in Table~\ref{tab:vocal-persona-infomation}.

\begin{table}[!htbp]
    \centering
    \caption{Paralinguistic and speaker characteristic information in GOAT-SLM.}\label{tab:vocal-persona-infomation}
    \resizebox{\textwidth}{!}{
    \begin{tabular}{lc}
        \hline
        \textbf{Dimension} & \textbf{Tags} \\\hline        
         Emotion & happy, sad,  angry, disgusted, fearful, surprised, neutral \\ %\hline
         Non-speech vocalization & cough, throat clearing, laughter, sneeze \\ \hline 
         Age & children, adults, elderly \\ %\hline
         Language/Dialect &  Mandarin, English, Cantonese, Shanghainese, Sichuanese, northeastern Mandarin, Henan dialect \\ \hline 
    \end{tabular}
    }
\end{table}

We carefully configure the training samples with different combinations of text queries (TQ), speech queries (SQ), text response (TR), and speech response (SR), as detailed in the corresponding data configuration Table~\ref{tab:data-config}. The hyperparameter setting and training costs of each stage are detailed in Table~\ref{tab:hyper-param}. This modular design of training data allows each stage to target specific capabilities while maintaining a coherent overall learning objective. 
\begin{table}[h]
\centering
\caption{Data configuration for training.}\label{tab:data-config}
\resizebox{\textwidth}{!}{
\begin{tabular}{lcccc}\hline
     \textbf{Stage}& \textbf{Sample} & \textbf{Num. Samples} &\textbf{Hours of Speech} & \textbf{Data Source}  \\\hline
     Stage 1 & <TQ, TR> & 115k & - & Dialogue \\ %\hline
     Stage 2-1 & <SQ, TR> & 73M & $\sim$170k& ASR/TTS    \\ %\hline
     Stage 2-2 & <SQ, TR> & 53M & $\sim$85k & ASR/TTS, Dialogue  \\%\hline
     Stage 3-1 & <SQ, TQ, SR, TR>& 56M & $\sim$150k& ASR/TTS \\%\hline
     Stage 3-2 & <SQ, SR, TR> & 5.7M & $\sim$20k & ASR/TTS, Dialogue \\\hline
\end{tabular}
}
\end{table}

\begin{table}[h]
\centering
\caption{Hyperparameters for all training stages.}
\label{tab:hyper-param}
\resizebox{\textwidth}{!}{
\begin{tabular}{lcccccccc}
\hline
     \textbf{Stage} & \textbf{Optimizer} & \textbf{Num. Warm. Steps} & \textbf{Scheduler} & \textbf{Learn. Rate} & \textbf{Num. GPU} & \textbf{Batch Size} & \textbf{Num. Epoch} & \textbf{GPU Hours} \\\hline
     Stage 1 & Adam & - & Cosine& 6e-5& 32$\times$A800-80G & 16 & 2 & 42.6 \\%\hline
     Stage 2-1 & AdamW & 1000 & Linear & 5e-5 & 48$~\times~$A800-80G &  1152 & 1 & 2305.2 \\%\hline
     Stage 2-2 & AdamW & 1000 & Linear & 2.5e-5 & 48$~\times~$A800-80G & 1152 & 1 & 2251.2 \\%\hline
     Stage 3-1 & AdamW & 12000 & Cosine & 5e-5 & 32$~\times~$A800-80G & 768 & 2 & 4566.4 \\%\hline
     Stage 3-2 & AdamW & - & Linear & 1e-5 & 24$~\times~$A800-80G & 240 & 2 & 1495.2 \\\hline
\end{tabular}
}

\end{table}

\begin{figure}
    \centering
    \includegraphics[width=1.0\linewidth]{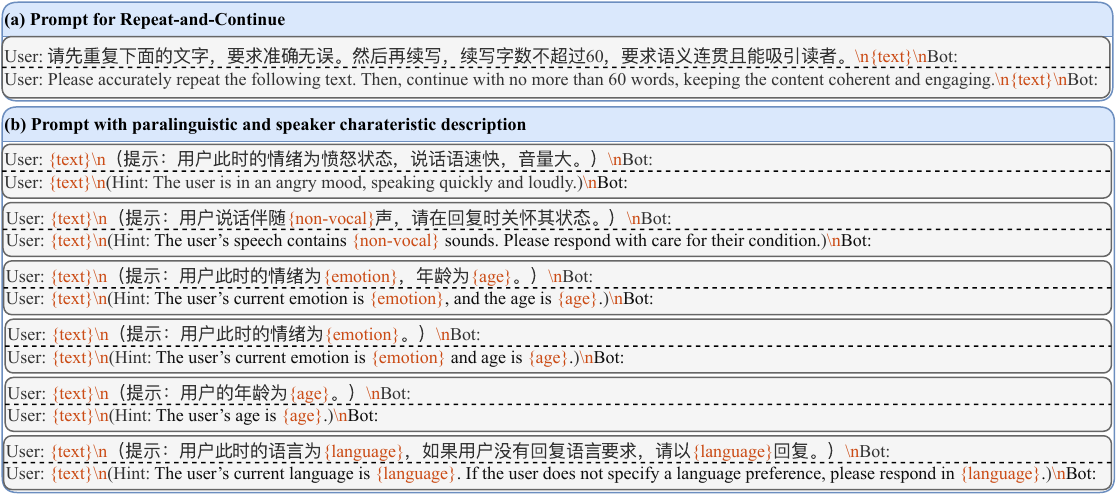}
    \caption{Prompts used to generate textual response during the construction of speech–text data pairs. The \textit{text}, \textit{non-vocal}, \textit{emotion}, \textit{language} and \textit{age} tags indicate the transcript, non-speech vocalizations, emotional state, language or dialect, and speaker’s age associated with the query audio, respectively.}
    \label{fig:prompts}
\end{figure}

\subsection{Stage 1: Instruction Tuning}
To enable the model to comprehend user instructions while simultaneously perceiving and adapting to paralinguistic and speaker-specific attributes, we perform supervised fine-tuning (SFT) on multi-turn dialogue data explicitly annotated with vocal characteristics.

Each user query in the training data consists of a natural instruction concatenated with a descriptive prompt that conveys vocal attributes such as age, dialect, emotion, non-speech vocal events (e.g., sighs, laughter), speaking rate, and volume. This setup provides the model with explicit cues about speaker traits while preserving the diversity of dialogue intents, which include knowledge-grounded QA, role-playing, task planning, and open-ended conversations. Dialogues are structured in multi-turn formats to reflect real-world interaction dynamics.

To enhance the model’s sensitivity to these vocal cues, we adopt a contrastive data construction strategy: for selected instructions, multiple versions are paired with varying vocal attribute descriptions, encouraging the model to produce distinguishable responses. This positive–negative contrast reinforces the alignment between vocal traits and output behavior.

We explore multiple prompting formats for integrating vocal attributes, including natural language descriptions and keyword-based labels, and use special tokens to support multi-attribute combinations while maintaining LLM fluency.

Model-generated responses are further filtered and manually refined to ensure naturalness, coherence, and context awareness, with careful attention to consistent integration of vocal attribute cues.

This stage establishes an alignment between explicit speaker-related attributes and adaptive textual responses, providing a foundation for paralinguistic-aware spoken language modeling in later stages.

\subsection{Stage 2: Speech–Text Alignment Training}
To align speech and text across both linguistic and paralinguistic dimensions, we adopt a self-distillation approach in which training targets are generated by prompting the LLM with structured transcripts similar to prior work~\cite{audiochatllama, blsp, wav2prompt, desta2}. We use a two-phase progressive strategy to ensure stability and effectiveness in this stage.

\textbf{Stage 2-1: Linguistic Alignment.}  
We begin by training on large-scale ASR datasets containing real-world spoken queries. The corresponding textual responses are generated using a “Repeat-and-Continue” prompting scheme, where the ASR transcript is provided as context (see Figure~\ref{fig:prompts}.a). This phase focuses on robust linguistic alignment without explicit paralinguistic signals. The parameters of the Think–Write module and the encoder of the Listen module are frozen, and only the projector in the Listen module is updated.

\textbf{Stage 2-2: Linguistic, Paralinguistic and Speaker Characteristic Alignment.}  
In this phase, speech queries consist of a mixture of real and synthesized utterances that reflect diverse paralinguistic and speaker characteristics, based on multi-turn dialogue inputs. 
The paralinguistic and speaker characteristic labels are obtained either through manual annotation or automatic classifiers. For each training sample, the LLM is prompted with transcripts and corresponding attribute descriptions (Figure~\ref{fig:prompts}.b) to generate target responses. All components of the Listen module are fine-tuned during this phase to enable alignment with both linguistic content and vocal attributes.

\subsection{Stage 3: High-Fidelity Expressive Speech Generation Training}

The final training stage aims to endow the model with the ability to generate high-quality, expressive speech responses that reflect rich paralinguistic and speaker-specific characteristics. This is achieved via a two-phase procedure, Stage 3-1 and Stage 3-2, similar in formulation to the GOAT-TTS framework~\cite{goat-tts}.

\textbf{Stage 3-1: Cold-start Speech Generation Training.} 
We initiate the Speak module with a data-driven joint optimization strategy using large-scale real-world datasets. For Mandarin and English, training inputs are structured as <TQ, TR, SR> triplets (Text Query, Text Response, Speech Response), while dialectal data adopt <SQ, TR, SR> triplets (Speech Query instead of text). This unified paradigm facilitates the mapping from linguistic content and latent attributes (e.g., emotion) to speech output while supporting dialect following. To further enhance robustness in long-form generation, speech segments from the same speaker are randomly concatenated to form samples up to 60 seconds.

\textbf{Stage 3-2: Attribute-Aware Refinement.}  
This sub-stage enhances the model’s capacity for stylistically adaptive speech generation. We first construct multi-dimensional speech prompt sets by collecting recordings from a professional voice actor, covering four core emotions (happiness, comfort, surprise, neutral) and three target listener demographics (children, adults, elderly). In parallel, natural conversational segments are extracted from dialect corpora and converted into target vocal styles using GOAT-TTS’s flow module to ensure timbre consistency.

We then construct training triplets by pairing filtered samples from Stage 2-2—categorized by emotion, age, and language—with the collected speech prompts. Using GOAT-TTS, the textual responses are converted into speech targets to form new <SQ, TR, SR> samples. Fine-tuning on this data enables the model to produce speech outputs that are emotionally expressive, age-aware, and dialect-sensitive in diverse conversational contexts.

\textbf{Training Configuration and Optimization.}  
During this phase, parameters of the Listen and Think modules are frozen, while training is focused on the Speak module. By preserving the Think module, the model retains robust text comprehension capabilities, allowing the Speak module to handle noisy or non-standard text inputs—such as emojis, line breaks, or special punctuation—generated by the Write module.

To support temporally coherent speech synthesis, we introduce a Multi-Token Prediction (MTP) mechanism. At each decoding step, the current output token embeddings are concatenated with the next-step input features, forming a fused latent representation that guides the generation of subsequent tokens. This mechanism, combined with the dual-modality head architecture, allows the Speak module to perform low-latency, streaming speech generation with a one-step delay behind the Write module. Cache management strategies are also employed to maintain efficiency and stability during multi-turn dialogue without requiring iterative fine-tuning.

Notably, we adopt a confidence-based automatic gradient masking strategy during training: high-confidence speech tokens receive full gradient updates, while low-confidence tokens are masked. This selective optimization significantly enhances pronunciation stability and overall speech fidelity—a technique also leveraged in GOAT-TTS.

\section{Evaluation}
To systematically assess GOAT-SLM’s capabilities, we evaluate the model using TELEVAL$^1$, a multi-dimensional benchmark suite designed to measure two key areas: (1) semantic intelligence and (2) paralinguistic and speaker characteristic-aware interaction.
The semantic intelligence evaluation includes multilingual question answering, dialectal QA, and multi-turn dialogue, targeting the model’s ability to understand, reason, and respond accurately in spoken interactions.
The paralinguistic and speaker characteristic-aware evaluation covers emotion-conditioned response generation, dialectal adaptation, age-aware interaction, and non-speech vocal response, focusing on the model’s ability to perceive vocal traits and produce adaptive spoken responses.
\footnotetext{$^1$https://github.com/Tele-AI/TELEVAL}

\subsection{Results of Semantic Intelligence}

To evaluate models' semantic intelligence, we first adopt an Audio Question Answering (AQA) approach to test their performance on general knowledge QA in both Chinese and English. As shown in Table~\ref{AQA_results}, no open-source model achieves consistently top performance across all datasets. However, MiniCPM-o 2.6~\cite{minicpm-o}, Qwen2.5-Omni~\cite{Qwen25-omni}, and Kimi-Audio~\cite{kimi-audio} each demonstrate strong performance on specific subsets. For GOAT-SLM, its capability in general AQA slightly declines due to the incorporation of paralinguistic and speaker characteristic awareness, yet it still remains at an average level overall.

\begin{table}[!htbp]
\caption{Results (\%) on the Common AQA task in Chinese and English.}
\label{AQA_results}
\centering
\resizebox{\textwidth}{!}{
    \begin{tabular}{lcccccccc}
    \hline
    \textbf{Model} & \textbf{LlamaQA-en} & \textbf{LlamaQA-zh} & \textbf{TriviaQA-en} & \textbf{TriviaQA-zh} & \textbf{WebQ-en} & \textbf{WebQ-zh} & \textbf{ChineseSimpleQA-zh} & \textbf{ChineseQuiz-zh} \\ 
    \hline
    GLM-4-Voice~\cite{glm4voice} & 67.67 & 53.00 & 34.89 & 27.00 & 37.00 & 34.62 & 14.47 & 47.09 \\
    MiniCPM-o~\cite{minicpm-o} 2.6 & 70.67 & 58.33 & 46.95 & 30.59 & 48.50 & 39.42 & 13.68 & 46.25 \\
    Baichuan-Omni-1.5~\cite{baichuan-omni15} & 69.33 & 58.00 & 34.89 & 29.75 & 42.98 & 39.32 & 15.74 & 51.09 \\
    SpeechGPT-2.0-preview~\cite{speechgpt2} & 0.00 & 36.33 & 0.12 & 13.62 & 0.00 & 20.33 & 4.16 & 27.12 \\
    Freeze-Omni~\cite{freeze-omni} & 66.00 & 57.67 & 37.87 & 23.78 & 41.95 & 35.60 & 14.48 & 49.76 \\
    Qwen2.5-Omni~\cite{Qwen25-omni} & 69.67 & \textbf{58.67} & 43.13 & 29.03 & 44.32 & 35.19 & 13.42 & 56.30 \\
    Kimi-Audio~\cite{kimi-audio} & 70.67 & 65.33 & 45.52 & \textbf{32.97} & 43.81 & 39.27 & 17.58 & 53.51 \\
    \hline
    GOAT-SLM & 72.33 & 52.67 & 37.51 & 28.43 & 39.73 & 35.14 & 14.47 & 48.43 \\ 
    \hline
    \end{tabular}
}
\end{table}

We further evaluate the model's capability to understand dialectal speech in the AQA task using audio inputs in five Chinese dialects: Cantonese, Henan dialect, Northeastern Mandarin, Sichuan dialect, and Shanghainese. The goal is to assess whether the models can still provide accurate answers when user questions are spoken in dialects, as compared to their performance on standard Mandarin in the ChineseQuiz-zh dataset. As shown in Table~\ref{dialect_AQA_results}, Baichuan-Omni-1.5, Qwen2.5-Omni, Kimi-Audio, and GOAT-SLM demonstrate some degree of dialect comprehension. However, their performance generally declines to varying extents compared to ChineseQuiz-zh. Among the dialects, Northeastern Mandarin yields the best performance across most models, likely due to its relatively small lexical divergence from Standard Mandarin. For more challenging dialects such as Cantonese and Shanghainese, only Baichuan-Omni-1.5, Qwen2.5-Omni, and GOAT-SLM achieve relatively strong performance. Notably, Baichuan-Omni-1.5 and Qwen2.5-Omni have been trained on a substantial amount of audio data that is more than 4 times larger than our training corpus, which may contribute to their robustness in handling dialectal variation.

\begin{table}[!htbp]
\caption{Results (\%) on the Dialect AQA task in Cantonese, Henan dialect, Northeastern Mandarin, Shanghainese, Sichuanese.}
\label{dialect_AQA_results}
\centering
\resizebox{\textwidth}{!}{
    \begin{tabular}{lcccccc}
    \hline
    \multirow{2}{*}{\textbf{Model}} & \multicolumn{5}{c}{\textbf{ChineseQuiz}} & \multirow{2}{*}{Average} \\ \cline{2-6}
     & \textbf{cantonese} & \textbf{henan\_dialect} & \textbf{northeastern\_mandarin} & \textbf{shanghainese} & \textbf{sichuanese} &  \\ \hline
    GLM-4-Voice & 0.61 & 9.93 & 37.40 & 3.87 & 13.35 & 13.13 \\
    MiniCPM-o 2.6 & 15.17 & 10.46 & 35.77 & 1.85 & 17.80 & 16.67 \\
    Baichuan-Omni-1.5 & 31.71 & 25.00 & 43.25 & 12.73 & 37.39 & 30.68 \\
    SpeechGPT-2.0-preview & 0.30 & 3.37 & 15.77 & 1.29 & 4.01 & 4.98 \\
    Freeze-Omni & 1.06 & 13.83 & 38.05 & 2.95 & 24.78 & 16.44 \\
    Qwen2.5-Omni & 48.10 & 34.75 & 46.99 & 24.72 & 44.81 & 40.54 \\
    Kimi-Audio & 17.91 & 24.65 & 42.76 & 4.24 & 35.91 & 25.71 \\
    \hline
    GOAT-SLM & 33.08 & 30.85 & 35.93 & 21.03 & 31.01 & 30.65 \\
    \hline
    \end{tabular}
}
\end{table}

The evaluation of the models’ ability to handle and retain information in multi-turn interactions uses a dataset of 150 constructed multi-turn dialogues. To prevent semantic drift caused by potentially uncontrollable model responses in intermediate turns, we design the dialogues such that the user provides information in the early turns, and the model is only required to acknowledge the input. The actual question is asked only in the final turn. As shown in Table~\ref{multiturn_results}, most models demonstrate multi-turn reasoning capabilities comparable to the text-based LLM counterparts. Notably, GOAT-SLM, despite not being trained with any multi-turn dialogue data, still performs competitively. This may be attributed to its well-aligned input embeddings under the current training paradigm.

\begin{table}[!htbp]
\centering
\caption{Results (\%) on the Multi-turn Dialogue task.}
\label{multiturn_results}
    \begin{tabular}{lc}
    \hline
    \textbf{Model} & \textbf{Multiturn\_memory-zh} \\
    \hline
    GLM-4-Voice & 80.00 \\
    MiniCPM-o 2.6 & 86.67 \\
    Baichuan-Omni-1.5 & 78.67 \\
    SpeechGPT-2.0-preview & 20.00 \\
    Freeze-Omni & 62.67 \\
    Qwen2.5-Omni & 88.67 \\
    \hline
    GOAT-SLM & 84.00 \\
    \hline
    \end{tabular}
\end{table}

\subsection{Results of Paralinguistic and Speaker Characteristic Awareness}
First, we evaluate the models’ ability to perceive and respond to paralinguistic and speaker characteristics on the dialect following task. Similar to the dialectal AQA task, we conduct experiments using the same five Chinese dialects. As shown in Table~\ref{dialect_follow_results}, open-source models that previously achieved strong performance in the dialectal AQA task (Table~\ref{dialect_AQA_results}) perform noticeably worse in this setting. This suggests that, while the models may comprehend dialectal content, they generally fail to recognize and naturally mirror the user’s dialectal style in their responses without extra instruction. GOAT-SLM continues to exhibit strong performance consistent with its results on the dialectal AQA task. This indicates that the model is not only capable of understanding dialectal input and performing question answering, but also able to generate appropriate dialectal responses in open-domain dialogue settings without requiring any explicit instructions.

\begin{table}[!htbp]
\centering
\caption{Results (\%) on the Dialect Following task.}
\label{dialect_follow_results}
\resizebox{\textwidth}{!}{
    \begin{tabular}{lcccccc}
    \hline
    \multirow{2}{*}{\textbf{Model}} & \multicolumn{5}{c}{\textbf{Chitchat}} & \multirow{2}{*}{\textbf{Average}} \\ \cline{2-6}
     & \textbf{cantonese} & \textbf{henan\_dialect} & \textbf{northeastern\_mandarin} & \textbf{shanghainese} & \textbf{sichuanese} &  \\ \hline
    GLM-4-Voice & 1.67 & 2.83 & 12.20 & 0.70 & 2.69 & 4.57 \\
    MiniCPM-o 2.6 & 8.42 & 9.44 & 21.27 & 2.67 & 10.33 & 10.98 \\
    Baichuan-Omni-1.5 & 6.40 & 7.06 & 11.48 & 2.74 & 8.67 & 7.38 \\
    SpeechGPT-2.0-preview & 0.70 & 4.40 & 13.11 & 1.08 & 4.00 & 5.17 \\
    Freeze-Omni & 0.70 & 5.81 & 10.94 & 1.29 & 9.42 & 5.72 \\
    Qwen2.5-Omni & 15.56 & 18.29 & 29.06 & 8.75 & 21.08 & 18.91 \\
    Kimi-Audio & 8.46 & 11.63 & 16.26 & 1.64 & 12.61 & 10.18 \\
    \hline
    GOAT-SLM & 71.38 & 32.42 & 53.93 & 45.20 & 47.61 & 50.73 \\ \hline
    \end{tabular}
}
\end{table}

Table~\ref{para_results} presents the evaluation results on three paralinguistic dimensions: emotion, non-speech vocal (NSV) signals, and speaker age. In the ESD-zh dataset, user utterances do not contain any explicit textual emotion cues, allowing us to assess whether models can infer the user's emotional state from vocal tone and respond appropriately. The Para\_mix300-zh test set extends standard AQA-style questions by combining four types of NSV signals, aiming to evaluate whether models can detect and respond to such cues. In the Age-zh dataset, queries are spoken using synthetic child or elderly voices, and the content of the questions is tailored to be age-appropriate.

When evaluating these three datasets, we focus not only on whether models can recognize or classify the paralinguistic features, but more importantly, on whether they can generate responses that are contextually and socially appropriate. The results show that most models are able to produce reasonably appropriate responses when the user's emotion or age is reflected in the input. However, all open-source models fail to effectively handle non-verbal vocal signals in their responses. Among them, Kimi-Audio demonstrates some ability to detect such signals, but still fails to generate contextually suitable replies. GOAT-SLM significantly outperforms all other models on both tasks.

\begin{table}[!htbp]
\centering
\caption{Results (\%) on the Paralinguistic task.}
\label{para_results}
    \begin{tabular}{lccc}
    \hline
    \textbf{Model} & \textbf{ESD-zh} & \textbf{Para\_mix300-zh} & \textbf{Age-zh} \\ \hline
    GLM-4-Voice & 35.55 & 1.89 & 27.81 \\
    MiniCPM-o 2.6 & 44.03 & 2.08 & 34.56 \\
    Baichuan-Omni-1.5 & 13.55 & 1.80 & 12.24 \\
    SpeechGPT-2.0-preview & 22.59 & 1.52 & 23.63 \\
    Freeze-Omni & 20.72 & 1.85 & 13.68 \\
    Qwen2.5-Omni & 44.83 & 2.19 & 42.51 \\
    Kimi-Audio & 53.17 & 9.19 & 22.77 \\
    \hline
    GOAT-SLM & 45.31 & 40.91 & 72.13 \\ \hline
    \end{tabular}
\end{table}

We also evaluate the quality of audio responses and the models' ability to generate speech. Following the setup in TELEVAL, we use the ESD-zh dataset to assess three aspects: DNSMOS, CER, and Emotion. As shown in Table~\ref{audio_results}, while models perform similarly in terms of DNSMOS scores, GOAT-SLM consistently outperforms all other open-source models across the remaining speech-related metrics. These results highlight the advantages of its model architecture, training paradigm and GOAT-TTS's capability for high-quality data construction.

\begin{table}[!htbp]
\centering
\caption{Results of Spoken Response Generation Evaluation. }
\label{audio_results}
    \begin{tabular}{lccc}
    \hline
    {\textbf{Model}} & {\textbf{CER $\downarrow$}} & {\textbf{DNSMOS $\uparrow$}} & {\textbf{Emotion $\uparrow$}} \\ \hline
    GLM-4-Voice & 6.58 & 3.46 & 31.66 \\
    MiniCPM-o 2.6 & 2.58 & 3.52 & 34.26 \\
    Baichuan-Omni-1.5 & 7.89 & 3.40 & 24.74 \\
    SpeechGPT-2.0-preview & 17.27 & 2.46 & 27.48 \\
    Freeze-Omni & 4.88 & 3.49 & 41.05 \\
    Qwen2.5-Omni & 1.69 & 3.47 & 52.59 \\
    Kimi-Audio & 3.84 & 3.38 & 45.48 \\
    \hline
    GOAT-SLM & 1.57 & 3.46 & 61.48 \\ \hline
    \end{tabular}
\end{table}

\begin{table}[!tbp]
\centering
\caption{Subjective Evaluation Results for Dialectal Spoken Response Generation.}
\resizebox{\textwidth}{!}{
\label{dialect_results}
    \begin{tabular}{lccccc}
    \hline
    \multirow{2}{*}{\textbf{Model}} & \multicolumn{5}{c}{\textbf{Dialectal Spoken Response (\%) $\uparrow$}} 
    \\ \cline{2-6} 
     & \textbf{cantonese} & \textbf{henan\_dialect} & \textbf{northeastern\_mandarin} & \textbf{shanghainese} & \textbf{sichuanese} \\ \hline
    GOAT-SLM & 97.4 & 99.6 & 57.2 & 99.0 & 92.6 \\ \hline
    \end{tabular}
}
\end{table}
In addition, we evaluate dialectal speech generation using the Chitchat-dialect dataset. In the subjective evaluation, 10 native speakers are recruited for each dialect to determine whether the model-generated speech belong to the target dialect. Since other baseline models lack dialect follow capabilities, their samples are excluded from the subjective assessment. As shown in Table~\ref{dialect_results}, despite underperformance in northeastern Mandarin, GOAT-SLM achieves superior performance (with consistency rates exceeding 90\%) in the remaining four dialects. Through analysis of the evaluation results, we find that even though the text generated by the Write module does not exhibit dialect features, the generated speech can still effectively acquire dialect-specific cues through the representations from the Listen module. This phenomenon validates the effectiveness of our proposed dual-modality head architecture design: while maintaining independent optimization of the text generation module and the speech generation module, the implicit interaction through deep representations successfully achieves effective transmission of dialect characteristics.

\section{Conclusion}
In this work, we presented GOAT-SLM, an end-to-end spoken language model designed to support natural and adaptive voice-based interaction. The model features a dual-modality head architecture that decouples linguistic modeling from speech generation, and a modular, progressive training strategy that integrates large-scale pretrained language and speech models.
Through targeted training across semantic, paralinguistic, and speaker-specific dimensions, GOAT-SLM is equipped to interpret and respond to both linguistic content and nuanced vocal cues. Evaluation results on the TELEVAL benchmark show that GOAT-SLM achieves balanced performance across multiple tasks and outperforms existing open-source models in several aspects of paralinguistic and speaker-aware interaction.
While GOAT-SLM demonstrates strong capabilities in perceiving non-linguistic speech signals, further research is needed to enhance its fine-grained paralinguistic reasoning and its adaptability to more diverse and dynamic interaction scenarios.

% For arXiv
% \bibliographystyle{plainnat}%{plainnat}%{IEEEtran}
% \bibliographystyle{abbrvnat}%{plainnat}%{IEEEtran}
% \bibliographystyle{unsrtnat}
\bibliographystyle{IEEEtran}
\bibliography{ref}

% \appendix
% \section{Data Construction}
% \section{Training details}

\end{document}